\documentclass[letterpaper, 10 pt, journal, twoside]{IEEEtran}

\usepackage{amsmath,amsfonts}
\usepackage{algorithmic}
\usepackage{algorithm}
\usepackage{array}
\usepackage[caption=false,font=normalsize,labelfont=sf,textfont=sf]{subfig}
\usepackage{textcomp}
\usepackage{stfloats}
\usepackage{url}
\usepackage{verbatim}
\usepackage{graphicx}
\usepackage{cite}
\usepackage{booktabs}
\usepackage{orcidlink}
\usepackage{bm}
\usepackage{multirow}
\usepackage{float}
\usepackage{amstext}
\usepackage{amsthm}
\usepackage{mathrsfs}
\usepackage{makecell}
\usepackage{colortbl}
\usepackage{hyperref}
\usepackage{amssymb}

\title{S-BEVLoc: BEV-based Self-supervised Framework for Large-scale LiDAR Global Localization}

\author{Chenghao Zhang$^*$, Lun Luo$^*$, Si-Yuan Cao, Xiaokai Bai, Yuncheng Jin, Zhu Yu, Beinan Yu, Yisen Wang, \\and Hui-Liang Shen, \emph{Senior Member, IEEE}
\thanks{Manuscript received: March, 21, 2025; Revised May, 23, 2025; Accepted July, 15, 2025.
This paper was recommended for publication by Editor Javier Civera upon evaluation of the Associate Editor and Reviewers' comments.
This work was supported in part by the National Key Research and Development Program of China under grant 2023YFB3209800, and in part by the Jinhua Science and Technology Bureau Project under grant 2023-1-118. \emph{(Corresponding author: Hui-Liang Shen)}}

\thanks{$^*$Chenghao Zhang and Lun Luo contribute equally to this work.}

\thanks{Chenghao Zhang, Lun Luo (project leader), Xiaokai Bai, Zhu Yu, and Yisen Wang are with the College of Information Science and Electronic Engineering, Zhejiang University, Hangzhou 310027, China (e-mail: zch00@zju.edu.cn, luolun@zju.edu.cn, shawnnnkb@zju.edu.cn, yu\_zhu@zju.edu.cn, wangyisen@zju.edu.cn).}

\thanks{Si-Yuan Cao is with the Ningbo Innovation Center, Zhejiang University, Ningbo 315100, China (e-mail: cao\_siyuan@zju.edu.cn).}

\thanks{Yuncheng Jin is with the College of Information Engineering, China Jiliang University, Hangzhou 310018, China (yunchengjin@cjlu.edu.cn).}

\thanks{Beinan Yu and Hui-Liang Shen are with the College of Information Science and Electronic Engineering, Zhejiang University, Hangzhou 310027, China, and also with the Jinhua Institute of Zhejiang University, Jinhua 321299, China. (e-mail:  mr\_vernon@hotmail.com, shenhl@zju.edu.cn).}

\thanks{Digital Object Identifier (DOI): see top of this page.}

}

\begin{document}

\markboth{IEEE Robotics and Automation Letters. Preprint Version. Accepted July, 2025}
{Zhang \MakeLowercase{\textit{et al.}}: S-BEVLoc: BEV-based Self-supervised Framework for Large-scale LiDAR Global Localization} 

\maketitle

\begin{abstract}
    LiDAR-based global localization is an essential component of simultaneous localization and mapping (SLAM), which helps loop closure and re-localization. Current approaches rely on ground-truth poses obtained from GPS or SLAM odometry to supervise network training. Despite the great success of these supervised approaches, substantial cost and effort are required for high-precision ground-truth pose acquisition. In this work, we propose S-BEVLoc, a novel self-supervised framework based on bird's-eye view (BEV) for LiDAR global localization, which eliminates the need for ground-truth poses and is highly scalable. We construct training triplets from single BEV images by leveraging the known geographic distances between keypoint-centered BEV patches. Convolutional neural network (CNN) is used to extract local features, and NetVLAD is employed to aggregate global descriptors. Moreover, we introduce SoftCos loss to enhance learning from the generated triplets. Experimental results on the large-scale KITTI and NCLT datasets show that S-BEVLoc achieves state-of-the-art performance in place recognition, loop closure, and global localization tasks, while offering scalability that would require extra effort for supervised approaches.
\end{abstract}


\section{Introduction}
\label{sec:intro}
\IEEEPARstart{G}{lobal} localization is a critical component for simultaneous localization and mapping (SLAM) systems. It serves two primary purposes: loop closure detection and re-localization~\cite{yin2024survey}. In loop closure detection, it can identify previously visited locations and correct accumulated odometry drift. In re-localization, it can re-establish the robot position within the map when encountering tracking failures. Hence, robust global localization is crucial for SLAM systems due to its critical role in ensuring accurate mapping and positioning. 

Most existing LiDAR global localization approaches \cite{komorowski2021egonn,cattaneo2022lcdnet,shi2024lcrnet,luo2024bevplace++} adopt a supervised learning pipeline. They follow a two-stage process comprising place recognition and pose estimation, and use ground-truth poses for supervision throughout the pipeline. 
In the place recognition stage, the point cloud closest to a query is retrieved through global descriptor matching. To train these global descriptors, the current supervised approaches commonly use position-based~\cite{komorowski2021egonn, cattaneo2022lcdnet,shi2024lcrnet, luo2024bevplace++} thresholds to define positive and negative sample pairs, thereby guiding the network to generate discriminative global descriptors. Position-based thresholding aligns with localization task requirements, but it depends heavily on positional accuracy. In the pose estimation stage, distinctive local features are extracted to establish correspondences between frames, followed by pose estimation using Random Sample Consensus (RANSAC). To enhance the distinctiveness of local features, these approaches \cite{komorowski2021egonn,vidanapathirana2022logg3d,shi2024lcrnet} generally need precise ground-truth poses to generate correspondence keypoints for training.

\begin{figure}[t]
    \centering
    \includegraphics[width=0.8\linewidth]{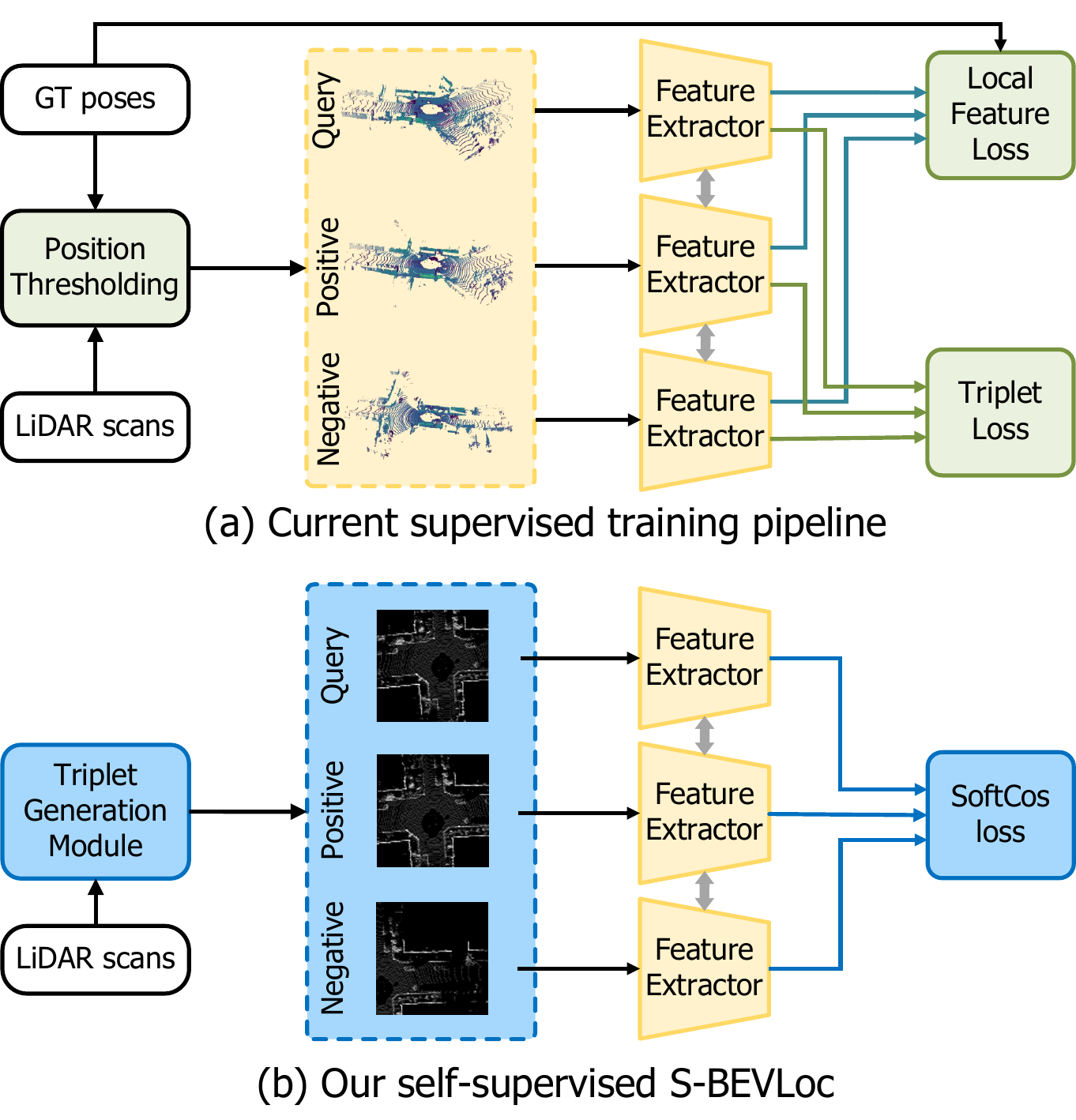}
	\vspace{-2mm}
    \caption{Comparison of current supervised training pipeline and our self-supervised method S-BEVLoc. (a) Supervised training uses ground-truth (GT) poses in position thresholding for training triplet generation and local feature supervision for local feature matching. (b) Our self-supervised method eliminates the need for ground-truth poses by generating the training triplets from single LiDAR scans, and performing feature matching directly using inherently distinctive local features.}
    \label{fig:overview}
	\vspace{-6mm}
\end{figure}

Although the supervised approaches have demonstrated satisfactory performance, their dependence on ground-truth poses introduces significant challenges. The acquisition of ground-truth poses necessitates precise sensor calibration (\emph{e.g.}, LiDAR, GPS, IMU) along with proper synchronization and interpolation, which are resource-intensive~\cite{geiger2012we,carlevaris2016university}. These requirements significantly limit the scalability of supervised approaches. In contrast, self-supervised approaches, which eliminate the reliance on ground-truth poses, remain underexplored despite their potential to address these limitations effectively.

This work proposes S-BEVLoc, a fully self-supervised LiDAR global localization framework based on bird's-eye view (BEV) images. S-BEVLoc is founded on the observation that triplets constructed from patches of a single BEV image can serve as hard triplets. Based on this insightful observation, we employ a simple yet effective strategy to construct training triplets using keypoint-centered BEV patches from single BEV images. This triplet construction strategy employs the known geographic distances of patches within a single BEV image for position thresholding, thus eliminating the need for ground-truth poses. In the pose estimation stage, inspired by BEVPlace++~\cite{luo2024bevplace++}, we employ a rotation equivariant module (REM) to extract rotation-equivariant and inherently distinctive local features, which can be directly used for feature matching, thereby eliminating the need for local feature supervision using ground-truth poses. 

For illustration, Fig.~\ref{fig:overview} compares our S-BEVLoc with the current supervised pipeline. S-BEVLoc takes single LiDAR scans as input and does not need ground-truth pose supervision. This frees us from the labor-intensive ground-truth pose generation and data pre-processing, and makes our method highly scalable. Moreover, we introduce SoftCos loss to enhance self-supervised learning from triplets derived from single LiDAR scans. We validate the effectiveness and scalability of S-BEVLoc using large-scale and long-term datasets. 
In summary, our contributions are as follows:
\begin{itemize}
    \item We propose a novel self-supervised method, S-BEVLoc, for LiDAR global localization. It takes single LiDAR scans as input and eliminates the need of ground-truth poses. To the best of our knowledge, this is the first attempt to achieve learning-based LiDAR global localization in a self-supervised manner. 
    \item We present an effective triplet generation strategy using single BEV images. We statistically validate that the constructed triplets are usually hard triplets which can boost the distinctiveness of the global descriptors. We also introduce SoftCos loss to enhance learning from these generated triplets.
    \item Extensive experiments validate that S-BEVLoc achieves comparable performance and robust generalization on large-scale and long-term datasets in a fully self-supervised manner and can offer high scalability.
\end{itemize}

\section{Related Work}
This section summarizes the recent advancements in LiDAR-based global localization, focusing on both handcrafted and learning-based approaches. Readers are referred to \cite{yin2024survey,zhang2023lidar,luo20243d} for comprehensive surveys.

\vspace{-2mm}
\subsection{Handcrafted Approaches}
Handcrafted approaches primarily exploit geometric priors from LiDAR data, eliminating the need for extensive training datasets. M2DP~\cite{he2016m2dp} projects points onto multiple 2D planes, generates density signatures, and computes global descriptors using singular vectors. BVMatch~\cite{luo2021bvmatch} employs BEV images to extract intensity and rotation-invariant bird's-eye view feature transform (BVFT) features, which are then aggregated into global descriptors using bag-of-words (BoW)~\cite{galvez2012bags}. BoW3D~\cite{cui2022bow3d} combines LinK3D~\cite{cui2024link3d} features with BoW for place recognition and 6-DoF pose correction in real-time loop closure. 3D-BBS~\cite{aoki20243d} enhances accuracy and processing speed by leveraging GPU acceleration and Branch-and-Bound (BnB)-based scan matching, requiring only a LiDAR scan roughly aligned with the gravity direction and a pre-built map. SOLiD~\cite{kim2024narrowing} presents a lightweight and rotationally robust solution for narrow FoV scenarios. A common limitation of these approaches is their limited scalability and robustness in dynamic environments, as they rely on handcrafted features that may not adapt well to diverse or changing scenarios. 

\vspace{-2mm}
\subsection{Learning-Based Approaches}
Learning-based approaches can be broadly categorized by their data representations to the (1) point cloud-based, (2) range image-based, and (3) BEV image-based ones. Point cloud-based approaches directly extract point-level and voxel-level features, making them sensitive to the distribution of LiDAR data and requiring substantial computational resources. To address this issue, several image-based approaches have been devised recently, leveraging mature techniques from computer vision. Typically, their pipelines involve first transforming LiDAR scans into image representations like range images~\cite{chen2022overlapnet,ma2022overlaptransformer,wu2021detailed} and BEV images~\cite{luo2023bevplace,luo2024bevplace++,jiang2024sg}, followed by the use of CNNs and Transformers to extract local features, and finally aggregating the features via NetVLAD~\cite{arandjelovic2016netvlad}.

\subsubsection{Point cloud-based approaches} LCDNet~\cite{cattaneo2022lcdnet} integrates unbalanced optimal transport to handle reverse loops and partial overlaps effectively. LoGG3D-Net~\cite{vidanapathirana2022logg3d} employs a local consistency loss to extract repeatable local features, which are then aggregated into global descriptors using second-order pooling and eigenvalue normalization.

\subsubsection{Range image-based approaches} OverlapNet~\cite{chen2022overlapnet} employs a Siamese network to predict overlaps and relative yaw angles between LiDAR scans. OverlapTransformer~\cite{ma2022overlaptransformer} designs a lightweight Transformer to generate yaw-angle-invariant global descriptors. CVTNet~\cite{ma2023cvtnet} aligns range and BEV image features using Transformers, creating robust yaw-angle-invariant descriptors.

\subsubsection{BEV image-based approaches} BEVPlace~\cite{luo2023bevplace} uses group convolutions for feature extraction and estimates query positions by mapping feature distances to geographic distances. BEVPlace++~\cite{luo2024bevplace++} enhances BEVPlace by introducing a rotation-equivariant module (REM) to extract robust features, facilitating zero-shot generalization to unseen environments. SG-LPR~\cite{jiang2024sg} adopts a semantic-guided framework with Swin Transformer and U-Net to improve feature extraction.

\subsubsection{Self-supervision for LiDAR} Although self-supervised approaches for LiDAR global localization remain underexplored, recent advancements~\cite{boulch2023also,SGDet3D,sautier2024bevcontrast} in semantic segmentation and object detection provide potential insights into self-supervised learning for LiDAR. ALSO~\cite{boulch2023also} leverages visibility information for self-supervised feature learning. BEVContrast~\cite{sautier2024bevcontrast} applies contrastive loss to 2D grid cell features, requiring known relative poses for pretraining.

\begin{figure*}[t]
    \centering
    \includegraphics[width=0.8\textwidth]{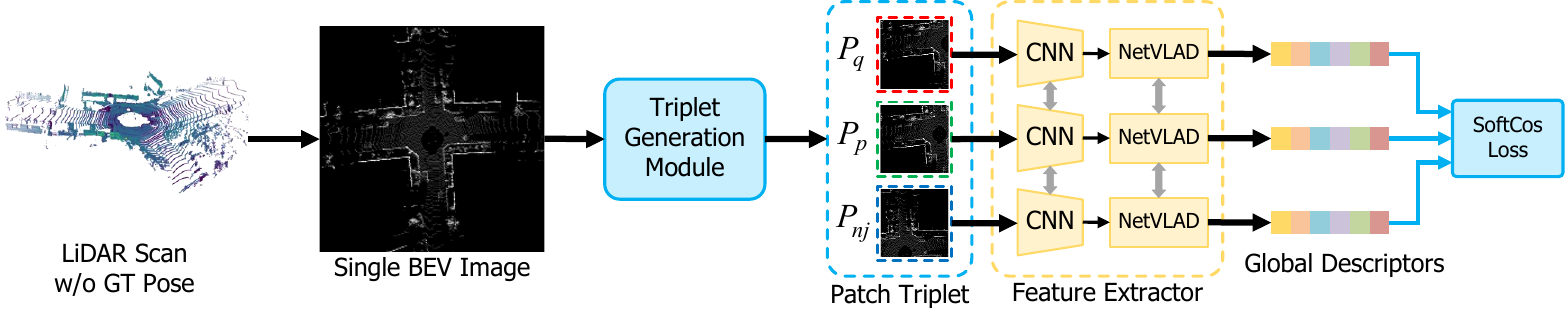}
	\vspace{-2mm}
    \caption{The overall training framework of S-BEVLoc. During training, a single LiDAR scan without ground-truth (GT) poses is first sampled and projected into BEV representation. The triplet generation module constructs a training patch triplet (see Fig.~\ref{fig:patch_module}). The patch triplet is used to train the feature extractor that comprises a CNN and a NetVLAD. The global descriptor generated by the feature extractor is supervised using our SoftCos loss.}
    \label{fig:framework}
	\vspace{-4mm}
\end{figure*}

\section{Method}
In this section, we first introduce the S-BEVLoc framework, and then elaborate our triplet generation strategy and SoftCos loss for self-supervised training. Finally, we present the global localization inference pipeline. 

\vspace{-2mm}
\subsection{Overview}
The overall self-supervised training framework of S-BEVLoc is depicted in Fig.~\ref{fig:framework}. S-BEVLoc uses BEV images~\cite{luo2021bvmatch} as an intermediate representation to perform place recognition and pose estimation. The training data consists of LiDAR scans without ground-truth pose information. 

During training, we first project a single LiDAR scan into BEV image via downsampling and gridding~\cite{luo2021bvmatch}, and then generate training triplets using the triplet generation module. We employ the CNN-based REM architecture from BEVPlace++~\cite{luo2024bevplace++} as the feature extractor, aiming to extract inherently distinctive and rotation-equivariant local features even without training. During inference, each query scan is first projected into BEV image, then a global descriptor is computed over the entire BEV image.

\vspace{-2mm}
\subsection{Triplet Generation Strategy}
The objective of triplet training is to learn a function $f(\cdot)$ that maps a BEV image $P$ to a discriminative global descriptor $\mathbf{V}=f(P)$, such that for any query $P^q$ with its positive and negative samples $P^+$ and $P^-$, the following relationship holds:
\begin{equation}
    \label{eq:objective}
    d(\mathbf{V}^q, \mathbf{V}^+) < d(\mathbf{V}^q, \mathbf{V}^-),
\end{equation}
where $\mathbf{V}^q=f(P^q)$, $\mathbf{V}^+=f(P^+)$, $\mathbf{V}^-=f(P^-)$, and $d(\cdot, \cdot)$ is a distance measure where Euclidean distance is commonly used. Hard mining~\cite{shrivastava2016training} is a commonly used strategy for identifying hard triplets, particularly those involving hard negatives that violate (\ref{eq:objective}) and introduce ambiguity in localization. By targeting these ambiguous samples, hard mining can enhance localization accuracy. Supervised approaches typically conduct hard mining based on ground-truth poses~\cite{luo2024bevplace++,vidanapathirana2022logg3d,komorowski2021egonn, cattaneo2022lcdnet}. For our self-supervised learning, we propose a triplet generation strategy that performs hard mining using single BEV images to construct hard triplets for training. 

\begin{figure}[t]
    \centering
    \includegraphics[width=0.85\linewidth]{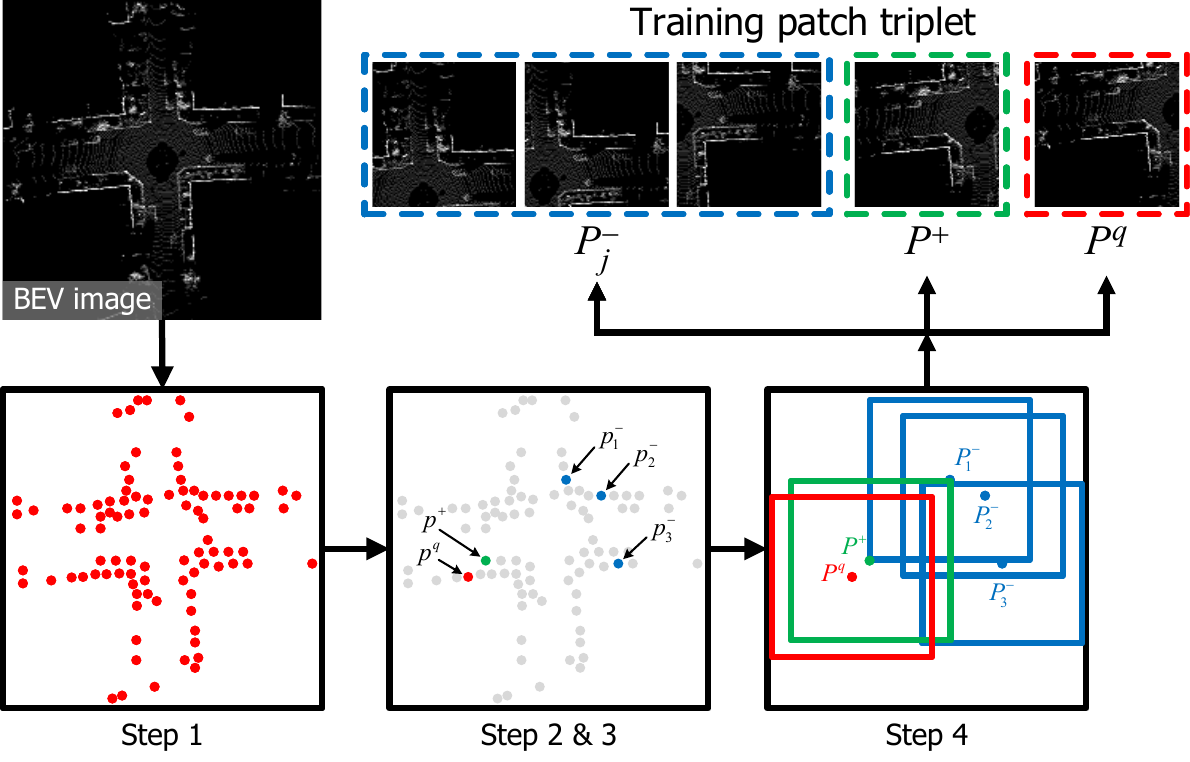}
	\vspace{-2mm}
    \caption{Illustration of the triplet generation module that implements our triplet generation strategy, consisting of four steps. Query, positive and negatives are denoted in red, green and blue, respectively.}
    \label{fig:patch_module}
	\vspace{-4mm}
\end{figure}

\textbf{Triplet Generation.}
As illustrated in Fig.~\ref{fig:patch_module}, the generation of training triplets from a LiDAR scan comprises 4 steps.

\emph{Step 1:} Given an input BEV image, we first detect the FAST~\cite{rosten2006machine} keypoints as the patch center point candidates $\mathcal{C}$. FAST is chosen for its computational efficiency and its ability to detect texture-rich corners such as building walls and road lamps. By using these keypoints as patch centers, we ensure that there is enough texture information in the patches and avoid generating patches that lack meaningful contents.

\emph{Step 2:} For a query point $p^q\in\mathcal{C}$, we calculate its geographic distance to the remaining points $\overline{\mathcal{C}}=\mathcal{C}\setminus\{p^q\}$. The geographic distance between two points is computed by scaling the pixel distance according to the downsampling factor, \emph{i.e.}, the resolution in BEV projection.

\emph{Step 3:} For the query point $p^q$, we generate its positive and negatives through position thresholding. Specifically, we randomly select a positive point ${p^+}$ from $\overline{\mathcal{C}}$ which has the geographic distance less than a distance threshold $D_\text{TH}$ (i.e., $\|p^+ - p^q\|<D_\text{TH}$), and select $m$ negative points $\{p^-_1,\dots,p^-_m\}$ with distances larger than $D_\text{TH}$. The query point, positive point and negative points together constitute a center point triplet $\mathcal{T}_\mathrm{center}=(p^q,p^+,\{p^-_j\})$. 

\emph{Step 4:} We take the points in $\mathcal{T}_\mathrm{center}$ as the centers, and crop the corresponding patches of size $r\times r$ from the full BEV image and apply zero-padding if out of bounds. These patches make up the training patch triplet $\mathcal{T}_\mathrm{patch}=(P^q,P^+,\{P^-_j\})$. We further apply data augmentations to prevent overfitting. 

\begin{figure}[t]
    \centering
    \includegraphics[width=\linewidth]{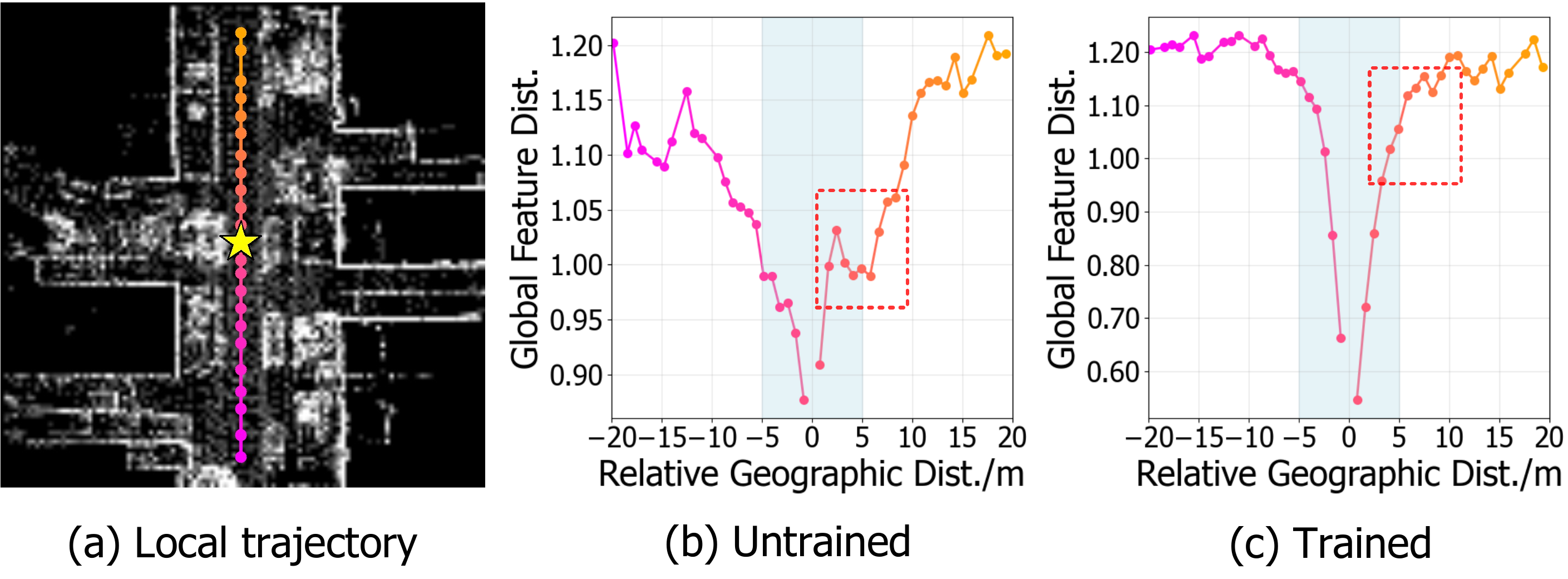}
	\vspace{-6mm}
    \caption{Effect of triplet training using our triplet generation strategy. (a) A local trajectory from KITTI sequence 00 consisting of adjacent samples to the query (denoted by a yellow star). (b) The trend of BEV global feature distance with respect to the geographic distance before training. (c) The trend after training using our triplet generation strategy. The light blue area indicates the range of positive samples after position thresholding with $D_\text{TH}=5~\text{m}$. We exclude the query itself for clarity. The sign of the geographic distance indicates chronological order. }
    \label{fig:observation}
	\vspace{-4mm}
\end{figure}

\textbf{Effect Illustration.}
We present an example to illustrate the effect of our triplet generation strategy on mitigating the ambiguity of hard negatives. Fig.~\ref{fig:observation}(a) depicts a local trajectory from KITTI sequence 00. We compute the Euclidean distance of global descriptors to obtain the BEV global feature distance, and plot its relationship with the geographic distance between a query sample (denoted by a star) and its adjacent samples. Fig.~\ref{fig:observation}(b) shows the relationship between global feature distance and geographic distance along the trajectory, using untrained global descriptors. Negatives near $D_\text{TH}=5$ meters have smaller global feature distances than positives, creating ambiguity that brings confusion to place retrieval. After training with our triplet generation strategy, as shown in Fig.~\ref{fig:observation}(c), the global feature distances of hard negatives become larger than those of positives, eliminating the ambiguity and demonstrating the effectiveness of our single BEV-based triplet generation strategy.

\vspace{-2mm}
\subsection{SoftCos Loss}
Our triplet generation strategy derives positive and negative samples from the same BEV image, requiring a loss function that effectively learns from these intra-image samples. This is essential for guiding the model in learning distinctive and robust global descriptors. 

The lazy triplet loss~\cite{uy2018pointnetvlad} is a widely used metric in supervised place recognition and global localization~\cite{cattaneo2022lcdnet,ma2022overlaptransformer,luo2023bevplace,luo2024bevplace++}. It is formulated as
\begin{equation}
    \label{eq:triplet}
    \mathcal{L}_\text{LazyTriplet}=\max_j([\delta^+ - \delta^-_j + \alpha]_+),
\end{equation}
where $\delta^+ = \|V^q-V^+\|$ and $\delta^-_j = \|V^q-V_j^-\|$ are the Euclidean distances between the query descriptor and its positive and $j$-th negative, while $\alpha$ denotes the fixed margin. 
To better accommodate our self-supervised training framework, we introduce SoftCos loss, defined as
\begin{equation}
    \mathcal{L}_\text{SoftCos}=\max_j\left\{\mathtt{Softplus}\left(s_j^- - s^+\right)\right\},
\end{equation}
where $s^+=\frac{V^q\cdot V^+}{\|V^q\| \|V^+\|}$ and $s_j^-=\frac{V^q\cdot V_j^-}{\|V^q\| \|V_j^-\|}$ are the positive and $j$-th negative cosine similarity with respect to the query descriptor, $\mathtt{Softplus}(x)=\tau\log\left(1+\exp\left(x/\tau\right)\right)$, and $\tau$ is a temperature hyperparameter  regulating the smoothness of the softplus function. Compared to the lazy triplet loss, our SoftCos loss differs in two key aspects: (1) The ReLU function ($[\cdot]_+$) and the margin ($\alpha$) are replaced by the softplus function, and (2) Euclidean distance is substituted with cosine similarity as the distance measure $d(\cdot, \cdot)$. 

\begin{figure}[t]
    \centering
    \includegraphics[width=0.7\linewidth]{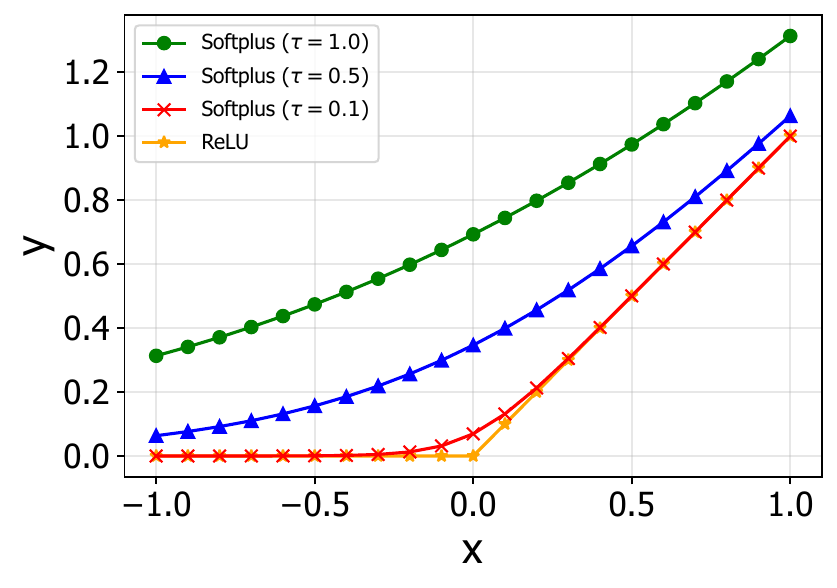}
	\vspace{-4mm}
    \caption{Comparison of ReLU and softplus with different temperatures $\tau$. }
    \label{fig:loss}
	\vspace{-4mm}
\end{figure}

\textbf{Explanation of Softplus.} For triplets that satisfy
\begin{equation}
    \label{eq:condition}
    \delta^+ - \delta^-_j + \alpha < 0,
\end{equation}
the ReLU function zeros out their gradients during backpropagation. This is too rigid for triplets generated by our triplet generation strategy, which may include noisy or ambiguous hard negatives. Consequently, valuable learning signals from such triplets satisfying (\ref{eq:condition}) may be discarded. In contrast, the softplus function preserves gradients for these triplets, ensuring they continue to contribute to learning more distinctive descriptors, as illustrated in Fig.~\ref{fig:loss}. Additionally, the softplus function implicitly introduces variable margin along the $x$-axis, where larger margins are applied to values of $x$ near zero, while smaller margins to values of $x$ far from zero. This adaptive margin mechanism is more effective than using a fixed margin for all $x$ values, as values of $x$ near zero correspond to beneficial hard triplets that provide useful learning signals. 

\textbf{Cosine Similarity.} 
For the L2-normalized global descriptors generated by NetVLAD (\emph{i.e.}, $\|V\|=1$), cosine similarity serves as a more natural and interpretable distance measure than Euclidean distance, as all discriminative information lies in direction rather than magnitude. Additionally, cosine similarity mitigates the curse of dimensionality~\cite{zhe2019directional}, leading to better generalization performance than Euclidean distance. 

\vspace{-2mm}
\subsection{Global Localization Inference Pipeline}
We perform global localization by first place recognition and then pose estimation, following previous works~\cite{luo2023bevplace,luo2024bevplace++}. In place recognition, we search the nearest neighbor of the query in a prebuilt BEV database as the retrieved place. In pose estimation, we estimate the global pose of the query by combining the pose of the retrieved place with the relative pose estimated through feature matching.

\textbf{Place Recognition.} A database of keyframe BEV images is first constructed from a pre-annotated map as $\mathcal{D}=\{(\mathbf{I}_i^\mathcal{D},\mathbf{T}_i^\mathcal{D},\mathbf{V}_i^\mathcal{D})|i=1,\dots,N\}$, where $\mathbf{T}_i^\mathcal{D}$ and $\mathbf{V}_i^\mathcal{D}$ are the associated global pose and global descriptor of keyframe BEV image $\mathbf{I}_i^\mathcal{D}$. Keyframes are selected by sampling scans at 1~m intervals, effectively reducing the database size by eliminating redundant observations of the same place, while ensuring comprehensive spatial coverage. Subsequently, place retrieval is carried out by seeking the nearest neighbor $\mathbf{V}^{q^*}$ of the query descriptor $\mathbf{V}^q$ by
\begin{equation}
    q^*=\mathop{\arg\min}\limits_{i=1,\dots,N}\|\mathbf{V}^q-\mathbf{V}_i^\mathcal{D}\|.
\end{equation}

\textbf{Pose Estimation.} After place retrieval, the FAST keypoints of both the query and the match BEV image are detected as the points of interest. The estimated relative pose $\mathbf{T}_\mathrm{rel}$ is obtained by applying RANSAC to match local features of these points. By combining the best match BEV image's ground-truth pose $\mathbf{T}_\mathrm{match}$, the global pose is obtained as
\begin{equation}
    \mathbf{T}_\mathrm{global}=\mathbf{T}_\mathrm{match}\mathbf{T}_\mathrm{rel}.
\end{equation}

\begin{table}[t]
    \belowrulesep=0pt
    \aboverulesep=0pt
    \footnotesize
    \renewcommand\arraystretch{0.9}
    \renewcommand\tabcolsep{4.5pt}
    \centering
    \caption{Recall at Top-1 on the KITTI dataset. The best and second-best ones are in bold and underlined, respectively.}
	\vspace{-2mm}
    \begin{tabular}{l|c|cccccc}
        \toprule[1.0pt]
        & Type & 00 & 02 & 05 & 06 & 08 & Mean\\
        \midrule
        M2DP~\cite{he2016m2dp} & / & 92.9 & 69.3 & 80.7 & 94.8 & 34.4 & 74.4 \\
        BoW3D~\cite{cui2022bow3d} & / & 71.4 & 15.5 & 58.7 & 91.8 & 57.0 & 58.9 \\
        LoGG3D-Net~\cite{vidanapathirana2022logg3d} & S & 99.6 & 96.1 & {97.5} & \textbf{100.0} & 93.5 & 97.3 \\
        LCDNet~\cite{cattaneo2022lcdnet} & S & \underline{99.9} & 97.7 & 95.3 & \textbf{100.0} & 94.4 & 97.4 \\
        CVTNet~\cite{ma2023cvtnet} & S & 98.7 & 87.1 & 93.5 & {97.8} & 83.7 & 92.1 \\
        BEVPlace~\cite{luo2023bevplace} & S & 99.7 & {98.1} & \underline{99.3} & \textbf{100.0} & 92.0 & 97.8 \\
        BEVPlace++~\cite{luo2024bevplace++} & S & \textbf{100.0} & \underline{99.3} & \textbf{100.0} & \textbf{100.0} & \underline{99.1} & \textbf{99.7} \\
        \midrule
        S-BEVLoc   & SS & {98.7} & \textbf{99.7} & {98.2} & \textbf{100.0} & \underline{99.1} & \underline{99.1}\\
        S-BEVLoc* & SS & {99.4} & {97.7} & {98.9} & \underline{99.6} & \textbf{99.7} & \underline{99.1}\\
        \bottomrule[1.0pt]
    \end{tabular}
    \label{tab:kitti}
	\vspace{-2mm}
\end{table}

\begin{table}[t]
	\vspace{-1.5mm}
    \belowrulesep=0pt
    \aboverulesep=0pt
    \footnotesize
    \renewcommand\arraystretch{0.9}
    \renewcommand\tabcolsep{5pt}
    \centering
    \caption{Recall at Top-1 on the rotated KITTI dataset. The best and second-best ones are in bold and underlined, respectively.}
	\vspace{-2mm}
    \begin{tabular}{l|c|cccccc}
        \toprule[1.0pt]
        & Type & 00 & 02 & 05 & 06 & 08 & Mean\\
        \midrule
        M2DP~\cite{he2016m2dp} & / & 92.9 & 69.3 & 80.7 & 94.8 & 34.4 & 74.4 \\
        BoW3D~\cite{cui2022bow3d} & / & 19.2 & 9.1 & 13.5 & 13.4 & 1.5 & 11.3 \\
        LCDNet~\cite{cattaneo2022lcdnet} & S & \textbf{99.7} & \underline{98.1} & 95.5 & \textbf{100.0} & 94.7 & {97.6} \\
        LoGG3D-Net~\cite{vidanapathirana2022logg3d} & S & {99.4} & 96.4 & 97.3 & \underline{99.6} & 92.0 & 96.9 \\
        CVTNet~\cite{ma2023cvtnet} & S & 98.7 & 87.4 & 93.3 & 98.5 & 85.8 & 92.7 \\
        BEVPlace~\cite{luo2023bevplace} & S & \underline{99.6} & 93.5 & \underline{98.9} & \textbf{100.0} & 92.0 & 96.8 \\
        BEVPlace++~\cite{luo2024bevplace++} & S & \textbf{99.7} & {97.1} & \underline{98.9} & \textbf{100.0} & {97.3} & \underline{98.6} \\
        \midrule
        S-BEVLoc   & SS & {99.3} & \textbf{98.4} & {98.2} & \textbf{100.0} & \underline{99.4} & \textbf{99.1}\\
        S-BEVLoc*  & SS  & {99.4} & {97.4} & \textbf{99.6} & \underline{99.6} & \textbf{99.7} & \textbf{99.1}\\
        \bottomrule[1.0pt]
    \end{tabular}
    \label{tab:rotkitti}
	\vspace{-2mm}
\end{table}

\begin{table*}[b]
	\vspace{-4mm}
	\belowrulesep=0pt
	\aboverulesep=0pt
	\footnotesize
	\renewcommand\arraystretch{0.9}
	\centering
	\caption{Recall at Top-1 on the NCLT dataset. Types are non-learning-based (/), supervised (S), and self-supervised (SS) approaches. The best and second-best ones are in bold and underlined, respectively.}
	\vspace{-2mm}
	\begin{tabular}{l|c|cccccc}
		\toprule[1.0pt]
		& Type & 2012-02-04 & 2012-03-17 & 2012-06-15 & 2012-09-28 & 2012-11-16 & 2013-02-23 \\
		\midrule
		M2DP~\cite{he2016m2dp} & / & 63.2 & 58.0 & 42.4 & 40.6 & 49.3 & 27.9 \\
		BoW3D~\cite{cui2022bow3d} & / & 14.9 & 10.7 & 6.5 & 5.0 & 5.2 & 7.5 \\
		LoGG3D-Net~\cite{vidanapathirana2022logg3d} & S & 69.9 & 19.6 & 11.0 & 8.7 & 10.9 & 25.6 \\
		LCDNet~\cite{cattaneo2022lcdnet} & S & 60.5 & 54.2 & 44.2 & 34.9 & 31.7 & 10.9 \\
		CVTNet~\cite{ma2023cvtnet} & S & 89.2 & {88.0} & 81.2 & 74.9 & 77.1 & 80.3 \\
		BEVPlace~\cite{luo2023bevplace} & S & {93.5} & {92.7} & {87.4} & {87.8} & {88.9} & {86.2} \\
		BEVPlace++~\cite{luo2024bevplace++} & S & {95.3} & {94.2} & {90.2} & {88.9} & {91.3} & {87.8} \\
		\midrule
		S-BEVLoc  & SS  & \underline{96.7} & \underline{95.4} & \underline{94.7} & \underline{90.8} & \underline{91.8} & \underline{90.4}\\
		S-BEVLoc* & SS & \textbf{96.9} & \textbf{95.6} & \textbf{94.9} & \textbf{91.7} & \textbf{92.0} & \textbf{91.3}\\
		\bottomrule[1.0pt]
	\end{tabular}
	\label{tab:nclt}
\end{table*}

\begin{table*}[t]
    \belowrulesep=0pt
    \aboverulesep=0pt
    \scriptsize
    \centering
    \renewcommand\arraystretch{0.96}
    \renewcommand\tabcolsep{0.75pt}
    \caption{Loop closure performance on KITTI and NCLT datasets. The best and second-best ones are in bold and underlined, respectively.}
	\vspace{-2mm}
    \begin{tabular}{l|ccccc|ccccc|ccccc|ccccc|ccccc|ccccc}
        \toprule[1.0pt]
        \multicolumn{1}{l|}{Sequence} 
        & \multicolumn{5}{c|}{KITTI 00} 
        & \multicolumn{5}{c|}{KITTI 02} 
        & \multicolumn{5}{c|}{KITTI 05} 
        & \multicolumn{5}{c|}{KITTI 06}    
        & \multicolumn{5}{c|}{KITTI 08} 
        & \multicolumn{5}{c}{\makecell{NCLT 2012-01-15}} 
        \\
        \midrule
        & AP & \makecell{max \\F1} & \makecell{max \\ R\%} & \makecell{$\hat{e}_t$ \\ (m)} & \makecell{$\hat{e}_r$\\$ (^\circ)$} 
        & AP & \makecell{max \\F1} & \makecell{max \\ R\%} & \makecell{$\hat{e}_t$ \\ (m)} & \makecell{$\hat{e}_r$\\$ (^\circ)$} 
        & AP & \makecell{max \\F1} & \makecell{max \\ R\%} & \makecell{$\hat{e}_t$ \\ (m)} & \makecell{$\hat{e}_r$\\$ (^\circ)$} 
        & AP & \makecell{max \\F1} & \makecell{max \\ R\%} & \makecell{$\hat{e}_t$ \\ (m)} & \makecell{$\hat{e}_r$\\$ (^\circ)$} 
        & AP & \makecell{max \\F1} & \makecell{max \\ R\%} & \makecell{$\hat{e}_t$ \\ (m)} & \makecell{$\hat{e}_r$\\$ (^\circ)$} 
        & AP & \makecell{max \\F1} & \makecell{max \\ R\%} & \makecell{$\hat{e}_t$ \\ (m)} & \makecell{$\hat{e}_r$\\$ (^\circ)$} 
        \\
        \midrule
        M2DP~\cite{he2016m2dp} 
        & 0.982 & 0.936 & 86.7 & - & -
        & 0.884 & 0.844 &  0.0 & - & -
        & 0.946 & 0.897 & 68.1 & - & -
        & 0.974 & 0.938 & 76.2 & - & - 
        & 0.081 & 0.162 & 0.0 & - & -
        & 0.783 & 0.695 & 4.8 & - & -
        \\
        BoW3D~\cite{cui2022bow3d} 
        & 0.979 & 0.897 & 46.5 & 0.54 & 1.20
        & 0.559 & 0.546 & 10.6 & 0.74 & 0.55
        & 0.957 & 0.857 & 47.8 & 0.69 & 0.72
        & 0.992 & 0.968 & 48.1 & 0.62 & 0.73 
        & 0.905 & 0.829 & 14.4 & 1.44 & 2.81
        & 0.000 & 0.000 & 0.0 & - & -
        \\
        LoGG3D-Net~\cite{vidanapathirana2022logg3d} 
        & 0.995 & {0.976} & 55.2 & - & -
        & \textbf{0.983} & \underline{0.927} & \underline{82.7} & - & -
        & \textbf{0.995} & {0.975} & 86.2 & - & -
        & {0.996} & 0.970 & 91.9 & - & - 
        & \underline{0.958} & \underline{0.929} & 2.7 & - & -
        & 0.679 & 0.592 & 1.0 & - & -
        \\
        LCDNet~\cite{cattaneo2022lcdnet} 
        & \underline{0.997} & 0.974 & {94.1} & \underline{0.10} & \underline{0.14}
        & {0.976} & 0.928 & \textbf{83.7} & \underline{0.65} & \textbf{0.44}
        & \underline{0.994} & 0.964 & {93.0} & \textbf{0.12} & \underline{0.17}
        & \textbf{0.999} & \underline{0.997} & \underline{99.6} & \textbf{0.11} & \underline{0.17}         
        & 0.952 & 0.918 & 12.2 & \textbf{0.21} & \textbf{0.47}
        & 0.633 & 0.342 & 0.0 & \underline{0.39} & \underline{1.20} 
        \\
        CVTNet~\cite{ma2023cvtnet} 
        & 0.994 & 0.965 & 84.8 & - & -
        & 0.931 & 0.898 & 64.6 & - & -
        & 0.975 & 0.933 & \textbf{96.2} & - & -
        & {0.996} & 0.981 & 96.2 & - & - 
        & 0.848 & 0.721 & 26.0 & - & -
        & 0.947 & {0.876} & 20.5 & - & -
        \\
        BEVPlace++~\cite{luo2024bevplace++} 
        &\textbf{0.999}& \textbf{0.995}& \textbf{98.4} & \textbf{0.08} & \textbf{0.11} 
        & \underline{0.977}& \textbf{0.934} & 70.0 & \textbf{0.38} & 0.70 
        & \underline{0.994} &\textbf{0.982} & \textbf{96.2} & \textbf{0.12} & \textbf{0.09} 
        & \textbf{0.999} & \textbf{0.999} & \textbf{100.0} & {0.18} & \textbf{0.08}  
        & \textbf{0.999} &\textbf{ 0.984} & \textbf{76.4} & \underline{0.35} & \underline{0.57} 
        & \underline{0.963} & \underline{0.912} & {24.9} & \textbf{0.34} & \textbf{1.09} 
        \\
        \midrule
        S-BEVLoc 
        & 0.993 & \underline{0.981} & \underline{96.1} & {0.11} & {0.16} 
        & 0.912 & 0.850 & 64.0 & {0.73} & \underline{0.52} 
        & 0.992 & 0.978 & \underline{95.2} & \underline{0.14} & \underline{0.17} 
        & \underline{0.998} & 0.985 & 97.0 & \underline{0.14} & {0.23} 
        & 0.919 & 0.843 & 52.8 & {0.63} & \underline{0.50} 
        & 0.947 & 0.885 & \underline{30.7} & {0.54} & {1.21} 
        \\
        S-BEVLoc* 
        & 0.992 & 0.974 & 92.3 & {0.12} & {0.16} 
        & 0.944 & 0.877 & 72.5 & {0.73} & \underline{0.52} 
        & 0.992 & \underline{0.979} & \underline{95.2} & {0.15} & {0.19} 
        & \textbf{0.999} & 0.993 & 98.5 & {0.18} & {0.26} 
        & 0.931 & 0.857 & \underline{66.7} & {0.61} & {0.52} 
        & \textbf{0.975} & \textbf{0.921} & \textbf{60.8} & {0.55} & {1.24} 
        \\
        \midrule
        \midrule
        \multicolumn{1}{l|}{} 
        & \multicolumn{5}{c|}{\makecell{NCLT 2012-02-04}} 
        & \multicolumn{5}{c|}{\makecell{NCLT 2012-03-17}}  
        & \multicolumn{5}{c|}{\makecell{NCLT 2012-06-15}} 
        & \multicolumn{5}{c|}{\makecell{NCLT 2012-09-28}} 
        & \multicolumn{5}{c|}{\makecell{NCLT 2012-11-16}} 
        & \multicolumn{5}{c}{\makecell{NCLT 2013-02-23}}   
        \\
        \midrule
        & AP & \makecell{max \\F1} & \makecell{max \\ R\%} & \makecell{$\hat{e}_t$ \\ (m)} & \makecell{$\hat{e}_r$\\$ (^\circ)$} 
        & AP & \makecell{max \\F1} & \makecell{max \\ R\%} & \makecell{$\hat{e}_t$ \\ (m)} & \makecell{$\hat{e}_r$\\$ (^\circ)$} 
        & AP & \makecell{max \\F1} & \makecell{max \\ R\%} & \makecell{$\hat{e}_t$ \\ (m)} & \makecell{$\hat{e}_r$\\$ (^\circ)$} 
        & AP & \makecell{max \\F1} & \makecell{max \\ R\%} & \makecell{$\hat{e}_t$ \\ (m)} & \makecell{$\hat{e}_r$\\$ (^\circ)$} 
        & AP & \makecell{max \\F1} & \makecell{max \\ R\%} & \makecell{$\hat{e}_t$ \\ (m)} & \makecell{$\hat{e}_r$\\$ (^\circ)$} 
        & AP & \makecell{max \\F1} & \makecell{max \\ R\%} & \makecell{$\hat{e}_t$ \\ (m)} & \makecell{$\hat{e}_r$\\$ (^\circ)$} 
        \\
        \midrule
        M2DP~\cite{he2016m2dp} 
        & 0.700 & 0.620 & 3.7 & - & -
        & 0.654 & 0.621 & 4.0 & - & - 
        & 0.666 & 0.617 & 1.9 & - & -
        & 0.676 & 0.602 & 4.2 & - & -
        & 0.281 & 0.380 & 0.0 & - & -
        & 0.700 & 0.656 & 1.3 & - & - 
        \\
        BoW3D~\cite{cui2022bow3d} 
        & 0.000 & 0.000 & 0.0 & - & -
        & 0.000 & 0.000 & 0.0 & - & - 
        & 0.024 & 0.102 & 0.0 & - & -
        & 0.000 & 0.000 & 0.0 & - & -
        & 0.000 & 0.000 & 0.0 & - & -
        & 0.000 & 0.000 & 0.0 & - & - 
        \\
        LoGG3D-Net~\cite{vidanapathirana2022logg3d} 
        & 0.575 & 0.517 & 0.6 & - & -
        & 0.570 & 0.530 & 1.4 & - & - 
        & 0.427 & 0.413 & 0.3 & - & -
        & 0.509 & 0.476 & 1.0 & - & -
        & 0.282 & 0.279 & 0.0 & - & -
        & 0.511 & 0.472 & 0.2 & - & - 
        \\
        LCDNet~\cite{cattaneo2022lcdnet} 
        & 0.621 & 0.362 & 0.0 & \underline{0.37} & {1.16}
        & 0.684 & 0.321 & 0.0 & \textbf{0.37} & \underline{1.26} 
        & 0.628 & 0.288 & 0.0 & \underline{0.50} & 1.30
        & 0.552 & 0.244 & 0.0 & \underline{0.44} & \underline{1.27}
        & 0.243 & 0.039 & 0.0 & \underline{0.47} & \underline{1.55}
        & 0.231 & 0.191 & 0.0 & \underline{0.52} & 1.68 
        \\
        CVTNet~\cite{ma2023cvtnet} 
        & 0.923 & {0.863} & {30.3} & - & -
        & 0.907 & {0.836} & {11.2} & - & - 
        & {0.937} & {0.869} & {36.8} & - & -
        & 0.920 & 0.840 & {19.7} & - & -
        & {0.784} & {0.719} & {8.1}  & - & -
        & 0.897 & 0.823 & {15.4} & - & - 
        \\
        BEVPlace++~\cite{luo2024bevplace++} 
        & \textbf{0.969} & \textbf{0.916} & \underline{34.5} & \textbf{0.36} & {1.19} 
        & \underline{0.935} & \underline{0.859} & \underline{31.2} & \underline{0.40} & \textbf{1.17} 
        & \textbf{0.955} & \underline{0.901} & {63.4} & \textbf{0.40} & {1.19} 
        & \underline{0.957} & \underline{0.894} & \textbf{45.3} & \textbf{0.40} & 1.61 
        & \underline{0.839} & \underline{0.733} & \underline{15.8} & \textbf{0.40} & \textbf{1.10} 
        & \underline{0.959} & \underline{0.887} & \underline{45.5} & \textbf{0.44} & {1.05} 
        \\		 
        \midrule
        S-BEVLoc 
        & 0.919 & 0.846 & 21.8 & {0.54} & \textbf{1.04} 
        & 0.896 & 0.832 & 26.3 & {0.57} & {1.29} 
        & \underline{0.946} & 0.889 & \underline{71.0} & {0.63} & \underline{1.04} 
        & 0.935 & 0.864 & 21.8 & {0.60} & \textbf{1.18} 
        & 0.743 & 0.713 & 2.8  & {0.85} & {2.05} 
        & 0.930 & 0.853 & 41.3 & {0.67} & \textbf{0.99} 
        \\
        S-BEVLoc* 
        & \underline{0.966} & \underline{0.913} & \textbf{41.7} & {0.54} & \underline{1.09} 
        & \textbf{0.956} & \textbf{0.889} & \textbf{53.9} & {0.56} & {1.31} 
        & \textbf{0.955} & \textbf{0.925} & \textbf{75.8} & {0.62} & \textbf{0.99} 
        & \textbf{0.970} & \textbf{0.922} & \underline{30.5} & {0.60} & \textbf{1.18} 
        & \textbf{0.883} & \textbf{0.826} & \textbf{29.3} & {0.88} & {1.98} 
        & \textbf{0.975} & \textbf{0.931} & \textbf{81.0} & {0.68} & \underline{1.04} 
        \\
        \bottomrule[1.0pt]
    \end{tabular}
    \label{tab:lc_ap_f1}
	\vspace{-2mm}
\end{table*}

\begin{table*}[t]
	\vspace{-2mm}
    \belowrulesep=0pt
    \aboverulesep=0pt
    \footnotesize
    \centering
    \renewcommand\tabcolsep{0.8pt}
    \caption{Global Localization Performance on NCLT dataset. The best and second-best ones are in bold and underlined, respectively.}
	\vspace{-2mm}
    \begin{tabular}{l|c|cccc|cccc|cccc|cccc|cccc|cccc}
        \toprule[1.0pt]
        & \multirow{3}{*}{Type} & \multicolumn{4}{c|}{2012-02-04} & \multicolumn{4}{c|}{2012-03-17} & \multicolumn{4}{c|}{2012-06-15} & \multicolumn{4}{c|}{2012-09-28} & \multicolumn{4}{c|}{2012-11-16} & \multicolumn{4}{c}{2013-02-23} \\
        \cmidrule{3-26}
        && \makecell{Recall$\uparrow$\\(\%)} & \makecell{SR$\uparrow$\\(\%)} & \makecell{$\bar{e}_t$$\downarrow$\\(m)} & \makecell{$\bar{e}_r$$\downarrow$\\($^\circ$)}
        & \makecell{Recall$\uparrow$\\(\%)} & \makecell{SR$\uparrow$\\(\%)} & \makecell{$\bar{e}_t$$\downarrow$\\(m)} & \makecell{$\bar{e}_r$$\downarrow$\\($^\circ$)}
        & \makecell{Recall$\uparrow$\\(\%)} & \makecell{SR$\uparrow$\\(\%)} & \makecell{$\bar{e}_t$$\downarrow$\\(m)} & \makecell{$\bar{e}_r$$\downarrow$\\($^\circ$)}
        & \makecell{Recall$\uparrow$\\(\%)} & \makecell{SR$\uparrow$\\(\%)} & \makecell{$\bar{e}_t$$\downarrow$\\(m)} & \makecell{$\bar{e}_r$$\downarrow$\\($^\circ$)}
        & \makecell{Recall$\uparrow$\\(\%)} & \makecell{SR$\uparrow$\\(\%)} & \makecell{$\bar{e}_t$$\downarrow$\\(m)} & \makecell{$\bar{e}_r$$\downarrow$\\($^\circ$)}
        & \makecell{Recall$\uparrow$\\(\%)} & \makecell{SR$\uparrow$\\(\%)} & \makecell{$\bar{e}_t$$\downarrow$\\(m)} & \makecell{$\bar{e}_r$$\downarrow$\\($^\circ$)}
        \\
        \midrule
        BoW3D~\cite{cui2022bow3d} & /
        & 14.9 & 3.8 & 1.11 & 2.08
        & 10.7 & 3.0 & 1.02 & 2.44
        &  6.5 & 1.1 & 1.23 & 2.62
        &  5.0 & 0.6 & 0.92 & 1.98
        &  5.2 & 0.3 & 1.27 & 2.36
        &  7.5 & 1.1 & 1.05 & 2.14
        \\
        LCDNet~\cite{cattaneo2022lcdnet} & S
        & 60.5 & 58.5 & \underline{0.37} & {1.15} 
        & 54.2 & 52.0 & \underline{0.37} & {1.26} 
        & 44.2 & 40.0 & \underline{0.49} & 1.28             
        & 34.9 & 32.2 & \textbf{0.44}    & \underline{1.27} 
        & 31.7 & 28.8 & \underline{0.47} & \textbf{1.54}    
        & 10.9 &  6.8 & \underline{0.50} & 1.62             
        \\
        BEVPlace++\cite{luo2024bevplace++}  & S
        & {95.3} & \textbf{95.6} & \textbf{0.32}    & \textbf{1.06}   
        & {94.2} & \textbf{95.1} & \textbf{0.33}    & \textbf{1.18}   
        & {90.2} & {90.9} & \textbf{0.42}    & \textbf{1.11}   
        & {88.9} & \textbf{89.8} & \underline{0.46} & \textbf{1.23}   
        & {91.3} & \textbf{90.2} & \textbf{0.44}    & {1.65}
        & {87.8} & \textbf{88.5} & \textbf{0.37}    & \textbf{1.05}   
        \\
        \midrule
        S-BEVLoc & SS
        & \underline{96.6} & {95.1} & {0.42} & \underline{1.14}
        & \underline{95.3} & {93.0} & {0.39} & \underline{1.24}
        & \underline{94.2} & \underline{91.8} & {0.51} & {1.24}
        & \underline{90.1} & {85.7} & {0.53} & {1.33}
        & \underline{91.5} & \underline{84.7} & {0.48} & {1.63}
        & \underline{90.0} & {86.5} & {0.52} & \underline{1.15}
        \\
        S-BEVLoc* & SS
        & \textbf{96.7} & \underline{95.3} & {0.43} & {1.17}
        & \textbf{95.8} & \underline{93.1} & {0.40} & \underline{1.24}
        & \textbf{94.7} & \textbf{92.1} & \underline{0.49} & \underline{1.22}
        & \textbf{92.0} & \underline{86.8} & {0.52} & {1.34}
        & \textbf{91.6} & \underline{84.7} & {0.48} & \underline{1.60}
        & \textbf{91.2} & \underline{87.0} & {0.52} & {1.16}
        \\
        \bottomrule[1.0pt]
    \end{tabular}
    \label{tab:global_localization}
	\vspace{-4mm}
\end{table*}

\vspace{-2mm}
\section{Experiments}
Our S-BEVLoc can directly use point clouds without ground-truth poses for training, whereas supervised approaches require considerable effort to first acquire ground-truth poses before training. To validate this scalability of S-BEVLoc, we present two models for evaluation: (1) S-BEVLoc, trained on KITTI sequence 00, and (2) S-BEVLoc*, a variant of S-BEVLoc trained on additional scans from KITTI sequence 11 to 21, which have no ground-truth poses. We evaluate the performance of S-BEVLoc and S-BEVLoc* in place recognition, loop closure, and global localization. We compare both models against open-sourced state-of-the-art (SOTA) approaches including (1) place recognition-only approaches M2DP~\cite{he2016m2dp}, LoGG3D-Net~\cite{vidanapathirana2022logg3d}, LCDNet~\cite{cattaneo2022lcdnet}, BEVPlace~\cite{luo2023bevplace}, and (2) place recognition and pose estimation approaches BoW3D~\cite{cui2022bow3d}, CVTNet~\cite{ma2023cvtnet}, and BEVPlace++~\cite{luo2024bevplace++}. We also conduct ablation study on loss function, point selection strategy, and sensitivity to key parameters.

\subsection{Datasets}
Following previous works~\cite{luo2023bevplace,luo2024bevplace++,jiang2024sg}, we conduct experiments on two large-scale LiDAR datasets: KITTI~\cite{geiger2012we} and NCLT~\cite{carlevaris2016university}. For each query frame, we regard its matched frame as true positive only if their geographic distance is less than 5 m. We present details of these datasets as follows:

\textbf{KITTI.} The KITTI Odometry dataset contains 22 sequences of point cloud data collected by a Velodyne HDL-64E LiDAR, with only sequence 00 to 10 providing ground-truth poses. We train S-BEVLoc using sequence 00, and S-BEVLoc* using sequence 00, 11 through 21. For evaluation, we select sequences that have ground-truth poses and contain large loops, including 00, 02, 05, 06 and 08. Each sequence is split into database frames and query frames for place recognition. Sequence 08 is considered the most challenging one among them since it consists of various reverse loops. 

\textbf{NCLT.} This dataset was collected on the University of Michigan's North Campus using a Velodyne HDL-32E LiDAR. The ground-truth poses are generated using LiDAR scan matching and high-accuracy RTK GPS. We select sequences with seasonal changes for evaluation, including 2012-01-15, 2012-02-04, 2012-03-17, 2012-06-15, 2012-09-28, 2012-11-16 and 2013-02-23. Last four sequences, 2012-06-15, 2012-09-28, 2012-11-16 and 2013-02-23, are considered challenging sequences because they contain stronger variations in weather, daytime, and season compared to the database.

\subsection{Implementation and Training Details}
We generate BEV images by cropping the point cloud to [-40~m, +40~m], and then downsampling with voxel grid filter of 0.4~m, resulting in a resolution of $200\times 200$. We set patch size $r=200$ to match the size of full BEV image. We incorporate an REM module with 8 rotations and a NetVLAD module with 64 clusters. For hyperparameters, we set number of negatives $m=10$, the temperature of softplus $\tau=0.1$, and the distance threshold $D_\text{TH}=5~\text{m}$. We train S-BEVLoc using the AdamW optimizer for 50 epochs with learning rate of $1\times 10^{-4}$ on an RTX 2080 Ti GPU.

\vspace{-6mm}
\subsection{Evaluation Metrics}
To evaluate the performance of S-BEVLoc, we adopt several widely-used metrics consistent with prior works~\cite{uy2018pointnetvlad,komorowski2021egonn,luo2023bevplace,luo2024bevplace++,shi2024lcrnet}.

For place recognition, we report the recall rate at Top-1, which is defined as the percentage of successfully recognized places. A place is considered successfully recognized if the retrieved Top-1 candidate is within $D_\text{TH}$ meters from the query.

For loop closure, we report the precision-recall (PR) curve, average precision (AP), maximum F1 score, maximum recall at 100\% precision (R\%), mean translation error, and mean rotation error. The PR curve is constructed by sweeping a decision threshold over the global descriptor distances. R\% is defined as the highest recall rate achieved while maintaining 100\% precision. The mean translation and rotation errors are computed over the successfully recognized places.

For global localization, we report the recall rate, success rate (SR), mean translation error, and mean rotation error. The SR is defined as the percentage of successfully localized queries. Following~\cite{shi2024lcrnet,luo2024bevplace++}, a localization is considered successful if the estimated global pose achieves translation and rotation errors below 2~m and 5$^\circ$, respectively.

\vspace{-2mm}
\subsection{Place Recognition}
We conduct experiments to evaluate place recognition performance in terms of robustness to view changes and generalization ability.

\textbf{Performance on KITTI.} Table~\ref{tab:kitti} presents recalls at Top-1 on the KITTI dataset. Both S-BEVLoc and S-BEVLoc* achieve comparable place recognition performance compared to supervised approaches. We also evaluate the robustness to view changes of both models on the rotated KITTI dataset, which is generated by applying random rotations around the z-axis to the LiDAR scans. As shown in Table~\ref{tab:rotkitti}, both S-BEVLoc and S-BEVLoc* achieve comparable performance to the supervised approaches.

\textbf{Generalization Ability.} To validate the generalization ability of S-BEVLoc, we evaluate the recall at Top-1 on the large-scale long-term NCLT dataset using the models trained on KITTI. We regard the sequence 2012-01-15 as the database and other sequences as queries. As shown in Table~\ref{tab:nclt}, S-BEVLoc outperforms all the supervised approaches, highlighting its superior generalization ability. Notably, S-BEVLoc* achieves an even higher recall rate, further demonstrating that scalability enhances generalization ability.

\begin{figure*}[t]
    \centering
    \includegraphics[width=0.9\textwidth]{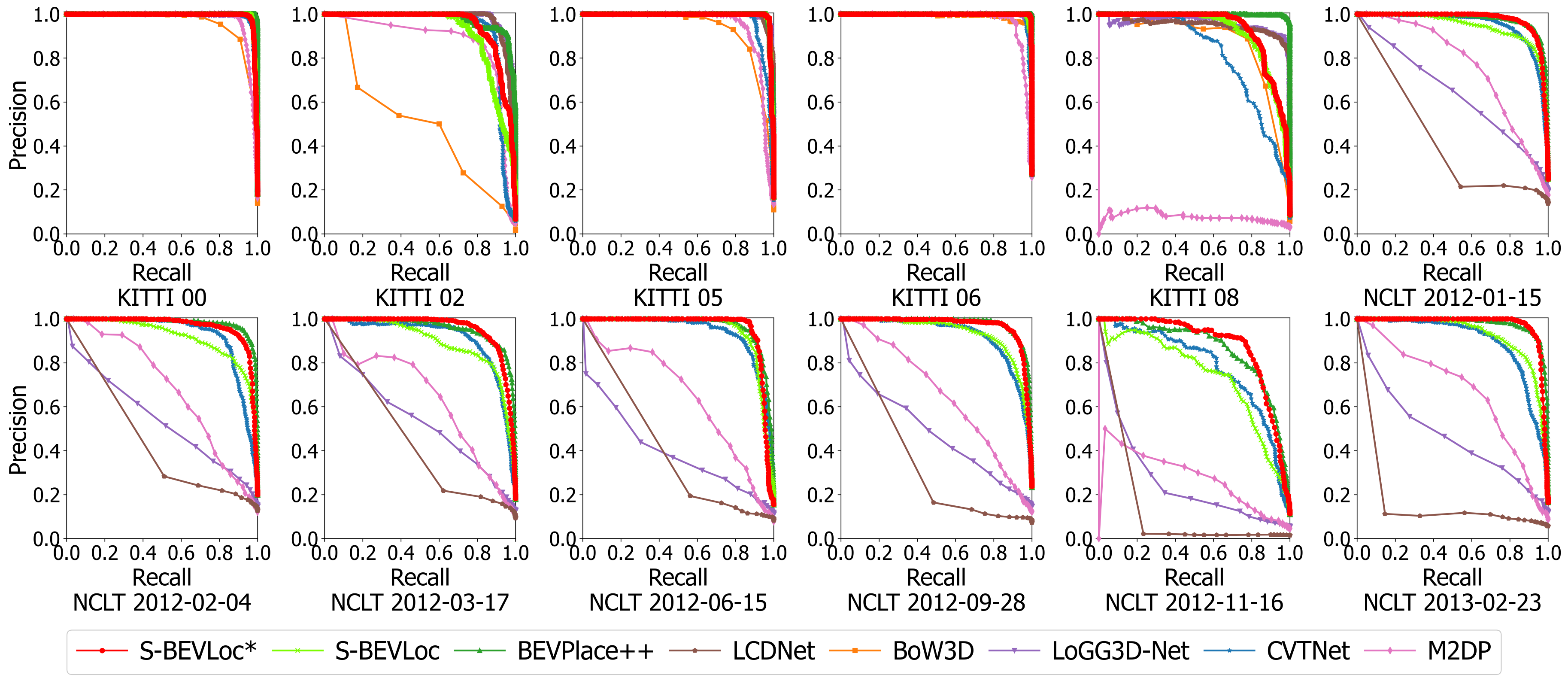}
	\vspace{-3mm}
    \caption{PR curves of different approaches on the KITTI and NCLT datasets.}
    \label{fig:pr_curve}
	\vspace{-4mm}
\end{figure*}

\vspace{-2mm}
\subsection{Loop Closure}
For loop closure, we perform place retrieval on previous frames, excluding the 100 most recent frames. 
Table~\ref{tab:lc_ap_f1} compares the loop closure performance of different approaches in terms of AP, maximum F1 score, R\%, mean translation error, and mean rotation error. S-BEVLoc achieves comparable performance against supervised approaches on all the sequences, and S-BEVLoc* generalizes even better than other approaches. Fig.~\ref{fig:pr_curve} presents the PR curves of different approaches on the KITTI and NCLT datasets. S-BEVLoc and S-BEVLoc* perform comparably to supervised approaches. 

\begin{table}[t]
	\vspace{-1.5mm}
    \belowrulesep=0pt
    \aboverulesep=0pt
    \footnotesize
    \centering
    \renewcommand\arraystretch{0.9}
    \renewcommand\tabcolsep{0.57pt}
    \caption{Recall at Top-1 of S-BEVLoc with different loss functions on challenging sequences. SoftCos-L2 denotes SoftCos with Euclidean distance as distance measure. Best ones are in bold.}
	\vspace{-2mm}
    \begin{tabular}{c|c|c|cccc}
        \toprule[1.0pt]
        \multirow{3}{*}{\makecell{Loss \\function}} & \multicolumn{1}{c|}{\multirow{2}{*}{KITTI}} & \multicolumn{1}{c|}{\multirow{2}{*}{\makecell{Rotated \\KITTI}}} & \multicolumn{4}{c}{\multirow{2}{*}{NCLT}} \\
        & & & \\
        \cmidrule{2-7}
        & 08 & 08 & 2012-06-15 & 2012-09-28 & 2012-11-16 & 2013-02-23 \\
        \midrule
        Lazy Triplet
        & 98.2 & 98.5 & 90.8 & 87.0 & 88.3 & 87.7
        \\
        SoftCos-L2
        & \textbf{99.1} & \textbf{99.4} & 91.2 & 87.7 & 89.0 & 88.7
        \\
        SoftCos
        & \textbf{99.1} & \textbf{99.4} & \textbf{94.7} & \textbf{90.8} & \textbf{91.8} & \textbf{90.4}
        \\
        \bottomrule[1.0pt]
    \end{tabular}
    \label{tab:loss}
	\vspace{-1.5mm}
\end{table}

\begin{table}[t]
	\vspace{-1.5mm}
    \belowrulesep=0pt
    \aboverulesep=0pt
    \footnotesize
    \centering
    \renewcommand\arraystretch{0.9}
    \renewcommand\tabcolsep{0.99pt}
    \caption{Recall at Top-1 of S-BEVLoc with different point selection strategies on challenging sequences. Best ones are in bold.}
	\vspace{-2mm}
    \begin{tabular}{c|c|c|cccc}
        \toprule[1.0pt]
        \multirow{3}{*}{\makecell{Point \\[-0.5ex]selection \\[-0.5ex]strategy}} & \multicolumn{1}{c|}{\multirow{2}{*}{KITTI}} & \multicolumn{1}{c|}{\multirow{2}{*}{\makecell{Rotated \\KITTI}}} & \multicolumn{4}{c}{\multirow{2}{*}{NCLT}} \\
        & & & \\
        \cmidrule{2-7}
        & 08 & 08 & 2012-06-15 & 2012-09-28 & 2012-11-16 & 2013-02-23 \\
        \midrule
        Random
        & 97.9 & 98.5 & 86.5 & 85.9 & 87.5 & 86.9
        \\
        Longitudinal
        & 97.9 & 98.8 & 92.2 & 90.2 & 89.1 & 89.3
        \\
        FAST
        & \textbf{99.1} & \textbf{99.4} & \textbf{94.7} & \textbf{90.8} & \textbf{91.8} & \textbf{90.4}
        \\
        \bottomrule[1.0pt]
    \end{tabular}
    \label{tab:strategy_cmp}
	\vspace{-2mm}
\end{table}

\begin{table}[t]
	\vspace{-1mm}
    \belowrulesep=0pt
    \aboverulesep=0pt
    \footnotesize
    \centering
    \renewcommand\arraystretch{0.9}
    \renewcommand\tabcolsep{1.5pt}
    \caption{Recall at Top-1 of S-BEVLoc with different patch sizes $r$ on challenging sequences. Best ones are in bold.}
	\vspace{-2mm}
    \begin{tabular}{c|c|c|cccc}
        \toprule[1.0pt]
        \multirow{3}{*}{\makecell{Patch \\size}} & \multicolumn{1}{c|}{\multirow{2}{*}{KITTI}} & \multicolumn{1}{c|}{\multirow{2}{*}{\makecell{Rotated \\KITTI}}} & \multicolumn{4}{c}{\multirow{2}{*}{NCLT}} \\
        & & & \\
        \cmidrule{2-7}
        & 08 & 08 & 2012-06-15 & 2012-09-28 & 2012-11-16 & 2013-02-23 \\
        \midrule
        50 &
        92.0 & 91.7 & 85.7 & 83.7 & 86.7 & 85.9
        \\
        100 &
        98.2 & 98.8 & 90.1 & 87.1 & 87.7 & 90.1
        \\
        200 &
        \textbf{99.1} & \textbf{99.4} & \textbf{94.7} & \textbf{90.8} & \textbf{91.8} & \textbf{90.4}
        \\
        \bottomrule[1.0pt]
    \end{tabular}
    \label{tab:patch_size}
	\vspace{-1.5mm}
\end{table}

\begin{table}[t]
	\vspace{-1.5mm}
    \belowrulesep=0pt
    \aboverulesep=0pt
    \footnotesize
    \centering
    \renewcommand\arraystretch{0.9}
    \renewcommand\tabcolsep{1.5pt}
    \caption{Recall at Top-1 of S-BEVLoc with different position thresholds on challenging sequences. Best ones are in bold.}
	\vspace{-2mm}
    \begin{tabular}{c|c|c|cccc}
        \toprule[1.0pt]
        \multirow{3}{*}{\makecell{Position \\[-0.5ex]threshold\\[-0.5ex]/m}} & \multicolumn{1}{c|}{\multirow{2}{*}{KITTI}} & \multicolumn{1}{c|}{\multirow{2}{*}{\makecell{Rotated \\KITTI}}} & \multicolumn{4}{c}{\multirow{2}{*}{NCLT}} \\
        & & & \\
        \cmidrule{2-7}
        & 08 & 08 & 2012-06-15 & 2012-09-28 & 2012-11-16 & 2013-02-23 \\
        \midrule
        5 & 
        \textbf{99.1} & \textbf{99.4} & \textbf{94.7} & \textbf{90.8} & \textbf{91.8} & \textbf{90.4}
        \\
        15 & 
        97.3 & 95.8 & 91.7 & 88.4 & 90.5 & 88.6
        \\
        25 & 
        97.0 & \textbf{99.4} & 91.8 & 87.0 & 90.0 & 87.7
        \\
        \bottomrule[1.0pt]
    \end{tabular}
    \label{tab:position_threshold}
	\vspace{-2mm}
\end{table}

\begin{table}[t]
	\vspace{-1mm}
    \belowrulesep=0pt
    \aboverulesep=0pt
    \footnotesize
    \centering
    \renewcommand\arraystretch{0.9}
    \renewcommand\tabcolsep{2pt}
    \captionsetup{labelfont={color=blue}, textfont={color=blue}}
    \caption{Recall at Top-1 of S-BEVLoc with different voxel sizes on challenging sequences. Best ones are in bold.}
	\vspace{-2mm}
    \begin{tabular}{c|c|c|cccc}
        \toprule[1.0pt]
        \multirow{3}{*}{\makecell{Voxel\\[-0.5ex]size\\[-0.5ex](m)}} & \multicolumn{1}{c|}{\multirow{2}{*}{KITTI}} & \multicolumn{1}{c|}{\multirow{2}{*}{\makecell{Rotated \\KITTI}}} & \multicolumn{4}{c}{\multirow{2}{*}{NCLT}} \\
        & & & \\
        \cmidrule{2-7}
        & 08 & 08 & 2012-06-15 & 2012-09-28 & 2012-11-16 & 2013-02-23 \\
        \midrule
        0.2 & {98.2} & {98.5} & {80.3} & {78.1} & {83.1} & {85.2}\\
        0.4 & \textbf{99.1} & \textbf{99.4} & \textbf{94.7} & \textbf{90.8} & \textbf{91.8} & \textbf{90.4}\\
        0.8 & {95.8} & {97.9} & {90.4} & {87.1} & {86.6} & {87.9}\\
        \bottomrule[1.0pt]
    \end{tabular}
    \label{tab:voxel_size}
	\vspace{-2mm}
\end{table}

\vspace{-2mm}
\subsection{Global Localization}
We evaluate global localization performance in terms of recall rate, SR, mean translation error, and mean rotation error. As shown in Table~\ref{tab:global_localization}, S-BEVLoc achieves comparable performance in terms of recall rate and SR, while remaining low translation error and rotation error. S-BEVLoc* shows higher recall rate and SR with lower localization error, highlighting the performance gains from easily attainable scalability.

\vspace{-2mm}
\subsection{Ablation Study}
\textbf{Ablation on Loss Function.} We compare the place recognition recalls of S-BEVLoc trained with three different loss functions: (1) lazy triplet loss, (2) SoftCos loss, and (3) SoftCos-L2 loss. SoftCos-L2 is a variant of SoftCos loss that uses Euclidean distance as the distance measure. Table~\ref{tab:loss} shows the Top-1 recall results on the challenging sequences. S-BEVLoc shows performance boost when trained with the SoftCos-L2 loss over the lazy triplet loss, and achieves the best performance while using the SoftCos loss. This validates the effectiveness of the Softplus function and demonstrates the generalization improvement brought by cosine similarity. 

\textbf{Ablation on Point Selection Strategy.} We compare our FAST keypoints-based point selection strategy with two vanilla ones: (1) randomly selecting points distributed across the BEV image, and (2) randomly selecting points along the vehicle's traveling direction, simulating the data acquisition process of LiDAR. Table~\ref{tab:strategy_cmp} illustrates the recall rates when using these different point selection strategies. 
Randomly selecting points distributed across the BEV image yields inferior recall results.
This is due to the generation of patches that lack meaningful content, which reduces the effectiveness of the training process. Limiting random point selection to the longitudinal axis improves generalization performance by reducing the selection of patches that lack meaningful content. Our FAST keypoints-based point selection strategy further mitigates the generation of such patches and achieves the best results. 

\textbf{Parameter Sensitivity.} We further investigate S-BEVLoc's sensitivity to hyperparameters including the patch size $r$, the distance threshold $D_\text{TH}$, and the voxel size in the BEV image generation. Table~\ref{tab:patch_size} presents the results of different $r$ values. A larger patch size leads to better performance because it enables the model to learn more consistent representations. Table~\ref{tab:position_threshold} shows the results of different $D_\text{TH}$ values. It is observed that S-BEVLoc performs the best when $D_\text{TH}=5~\text{m}$, which aligns with the defined positive threshold. Table~\ref{tab:voxel_size} presents the results of different voxel sizes in generating the BEV images. S-BEVLoc performs the best with voxel size of 0.4~m. A smaller voxel size leads to detail-rich BEV images but also introduces more noise, increasing the risk of overfitting, while a larger one produces overly compressed BEV images that lose critical details and degrade performance.

\vspace{-6mm}
\subsection{Running Time}
We test the running time of S-BEVLoc on a desktop equipped with an RTX 2080 Ti GPU and an Intel 3.70 GHz i5-12600KF CPU. For global localization, it takes about 23.2 ms for pairwise descriptor extraction (11.6 ms each), 0.121 ms for place retrieval, and 15.5 ms for pose estimation, achieving an average frequency of over 20 Hz. Considering that the frequency of LiDAR scans is usually 10 Hz, S-BEVLoc is suitable for real-time deployment. 

\section{Conclusions}
In this paper, we have proposed S-BEVLoc, a novel self-supervised framework based on BEV for LiDAR global localization. S-BEVLoc eliminates the dependence on ground-truth poses in both stages of LiDAR global localization. In the place recognition stage, S-BEVLoc leverages the known geographic distances between keypoint-centered BEV patches to construct training triplets from single BEV images. In the pose estimation stage, S-BEVLoc directly uses the inherently distinctive local features from CNN-based feature extractor for feature matching. Additionally, we introduced SoftCos loss to enhance self-supervised learning by employing the softplus function to retain valuable learning signals and cosine similarity to alleviate the curse of dimensionality. The experimental results show that S-BEVLoc achieves comparable performance and superior generalization ability compared to supervised approaches. In our future work, we will explore novel representations and network architectures to improve robustness against slope variations.

\bibliographystyle{IEEEtran}
\bibliography{ref}

\end{document}